\documentclass[10pt,twoside]{article}

\usepackage[utf8]{inputenc}
\usepackage[T1]{fontenc}
\usepackage{graphicx}
\usepackage{amsmath}

\usepackage{url}
\usepackage{xcolor}

\usepackage{caption}
\usepackage{subcaption}

\usepackage{siunitx}
\usepackage{taln2023}
\usepackage[french]{babel} 

\title{Comparative Analysis of Extrinsic Factors for NER in French}

\author{
Grace Yang\quad 
Zhiyi Li\quad 
Yadong Liu\quad 
Jungyeul Park\\
{\small Department of Linguistics, The University of British Columbia, Vancouver, BC V6T 1Z4, Canada \\ 
\texttt{\{grace04y,lizhiy1\}@student.ubc.ca, \{yadong.liu,jungyeul.park\}@ubc.ca}}
}

\begin{document}
\maketitle

\resume{
La reconnaissance d'entités nommées (REN) est une tâche cruciale qui vise à identifier des informations structurées, souvent remplies de termes techniques complexes et d'un degré élevé de variabilité.
REN précis et fiable peut faciliter l'extraction et l'analyse d'informations importantes. Cependant, le NER pour les autres langues que l'anglais est difficile en raison de la disponibilité limitée des données, car l'expertise, le temps et les dépenses élevés sont nécessaires pour annoter ses données. Dans cet article, en utilisant les données limitées, nous explorons divers facteurs extrinsèques, notamment la structure du modèle, le schéma d'annotation de corpus et les techniques d'augmentation des données pour améliorer les performances d'un modèle REN pour le français.
Nos expériences démontrent que ces approches peuvent améliorer considérablement le score F1 du modèle de 73,74 à 77,55. Nos résultats suggèrent que la prise en compte de différents facteurs extrinsèques et la combinaison de ces techniques est une approche prometteuse pour améliorer les performances du REN lorsque la taille des données est limitée.
}

\abstract{Benchmarking extrinsic factors for French NER.}{
Named entity recognition (NER) is a crucial task that aims to identify structured information, which is often replete with complex, technical terms and a high degree of variability. 
Accurate and reliable NER can facilitate the extraction and analysis of important information. However, NER for other than English is challenging due to limited data availability, as the high expertise, time, and expenses are required to annotate its data. In this paper, by using the limited data, we explore various factors including model structure, corpus annotation scheme and data augmentation techniques to improve the performance of a NER model for French. 
Our experiments demonstrate that these approaches can significantly improve the model's F1 score from original CRF score of 62.41 to 79.39. Our findings suggest that considering different extrinsic factors and combining these techniques is a promising approach for improving NER performance where the size of data is limited.\\
}
\motsClefs
  {Reconaissance d'entités nommées, augmentation de données, facteurs extrinsèques, REN français, extraction d’information}
  {Named Entity Recognition, Data Augmentation, Extrinsic factors, French NER, information extraction}

\section{Introduction}
Named entities are phrases that contain the names of persons, organizations and locations \cite{tjongkimsang-demeulder:2003:CONLL}. 
The task of named-entity recognition (NER) seeks to identify elements into predefined categories such as the names of persons (PER), locations (LOC), organizations (ORG), etc. 
The following example\footnote{The example excerpted from \url{https://github.com/EuropeanaNewspapers/ner-corpora}} is from Europeana Newspapers French NER data: \textit{M.}/\texttt{O} \textit{Brandi}/\texttt{I-PER} \textit{,}/\texttt{O}  \textit{professeur}/\texttt{O} \textit{au}/\texttt{O} \textit{lycée}/\texttt{I-ORG} \textit{de}/\texttt{I-ORG} \textit{Saint-Brieuc}/\texttt{I-ORG} (`{Mr. Brandi, teacher at high school of Saint-Brieuc}'). 

French Named entity recognition is essential. It is one of the most widely spoken languages in the world and improving the NER performance in French could help improve the acccuracy of various natural language processing tasks for a significant population of people. Accurately identifying named entities in French could be essential for tasks such as sentiment analysis, topic modeling and information extraction; thus, improving NER methods could have significant practical applications across a wide range of industries and research fields. 

However, French NER is less explored comparing to English one. English has received more attention in research and development for NER models. For French NER, it has less training data, pre-trained embeddings, data augmentation techniques and NER packages available comparing to English. This can make it more challenging to develop accurate and robust NER models for French.

The primary objective of this research paper is to examine the impact of extrinsic factors on the performance of French NER systems. To achieve this, we have implemented and evaluated a sequence labeling algorithm on a dataset of French text that has been labeled using various annotation schemes. In addition to this, we have explored the potential of data augmentation techniques to increase the size of our training dataset and enhance the performance of our NER models. By investigating the effects of different annotation schemes and data augmentation methods on NER performance, we aim to provide insights into how extrinsic factors can influence the accuracy and robustness of French NER systems. The findings of this research can be useful for improving the development of French NER models and enhancing their performance in real-world applications.

\section{Two extrinsic factors for French NER}

\subsection{Different annotation schemes}
Annotation representation (AR), a method of annotating each token with a predefined type in a given text, is widely used in sequence labelling tasks such as NER. The primary purpose of AR is to assign each token a tag which represents its position within a named entity and the category which it belongs to. The position of the token relative to its position in the named entity is marked by a single letter. 

In this study, three annotation schemes are used to investigate their influence on the task of French NER: 
(i) {IO:} The simplest annotation scheme. Each token that belongs to a named entity is represented with the {I}nside tag while tokens not belonging to a named entity are represented with the {O}utside tag. As this scheme does not mark the beginning or end of a named entity, it is unable to distinguish between consecutive entities. 
(ii) {BIO:} The first token of a named entity or a single-token named entity is assigned the {B}egin tag, while subsequent tokens in the same entity are assigned the {I}nside tag. Tokens not belonging to a named entity are tagged as {O}utside. 
(iii) {BIOES:} Increases the amount of information used to represent the boundaries of named entities. In addition to marking tokens that are at the {B}eginning, {I}nside, {E}nd, and {O}utside of a named entity, the {S}ingle tag is used to represent a named entity that consists of only one token. 

BIOES potentially performs best out of the three schemes, as the inclusion of additional tags may help to better distinguish between the boundaries of entities. This is because 'E-' and 'S-' tags can help better distinguish between last word of a named entity and singleton entities, which is particularly helpful when the language has complex morphologies and entities are long and complex. \citet{ratinov-roth:2009:CoNLL} concluded that the choice of annotation scheme has an impact on system performance, with BIOES outperforming the simpler BIO on English datasets. However, the BIOES also increases the number of tags in the dataset, which may affect the performance of the model. Therefore, it is important to weigh the benefits of additional tags against the potential complexity added to the model.

\begin{figure}
\centering
     \centering
     \begin{subfigure}[b]{0.45\textwidth}
         \centering
\scriptsize{
\begin{tabular}{ c c c c }
TOKEN & IO & BIO & BIOES \\  \hline
M. & O & O & O \\
Brandi & I-PER & \textbf{B}-PER & \textbf{S}-PER \\
, & O & O & O \\
Professeur & O & O & O \\
au & O & O & O \\
lycée & I-ORG & B-ORG & B-ORG \\
de & I-ORG & I-ORG & I-ORG \\
Saint-Brieuc & I-ORG & I-ORG & \textbf{E}-ORG \\
\end{tabular}
}         \caption{Example of IO,  BIO and BIOES}
         \label{figure-annotation}
\end{subfigure}
\begin{subfigure}[b]{0.45\textwidth}
\centering
\scriptsize{
\begin{tabular}{ c c c c }
LABEL & Original & LWTR & SIS \\  \hline
O & M. & M. & M. \\
S-PER  & Brandi & Louis & Brandi \\
O & , & , & , \\
O & Professeur & Professeur & Professeur \\
O & au & au & au \\
B-ORG & lycée & Château & de \\
I-ORG & de & de & Saint-Brieuc \\
E-ORG & Saint-Brieuc & Versailles & lycée \\
\end{tabular}
}
\caption{Example of DA}
         \label{figure-da}
     \end{subfigure}
\caption{Example of different annotation schemes and data augmentation}
\label{fig1}
\end{figure}

\subsection{Data augmentation}
Data augmentation is a commonly used technique in Natural Language Processing (NLP) to improve the performance of machine learning models. In NER, data augmentation expands the training set by modifying existing training instances without changing their labels. Models trained using data augmentation are thus more robust to variations in language and context \cite{wei-zou-2019-eda}. This positive effect occurs especially when datasets are small, as there is an increased risk of altering the ground-truth label or generating invalid instances on large datasets . In this paper, we used two data augmentation techniques: label-wise token replacement (LWTR) and shuffle within segments (SIS) from \cite{dai-adel-2020-analysis}. 

Label-wise token replacement involves randomly replacing a token of the same label based on label-wise token distribution built from a training set. In details, we first identify the named entities in the training data and their corresponding labels. We then randomly select a token from the named entity and replace it with another token that has the same label. For example, if the named entity "Paris" is labeled as a location, we could replace it with another location such as "London". It is important to note that it could also introduce noise to the dataset, especially as some of the replacement tokens might not be semantically or contextually similar to the original tokens. This method helps to create more varied examples within each label. 

Shuffle within segments splits the token sequence into segments of the same label and shuffles them based on a binomial distribution. In details, to apply the technique, we first divide the text into segments based on the presence of named entities. Then we randomly shuffle the tokens within each segment while keeping the segments themselves in original order. This method can help to create examples with a different order of the tokens, which can also increase the diversity of the examples. Though it may create semantically or contexually incoherent sequences, the varied and diverse examples can help prevent overfitting and improve the robustness and generalizability of the model. 



\section{Experiments and results}

We used French NER data as provided by Europeana Libraries. This data is based on OCRed and
manually annotated historical newspapers from the National Library of France and are originally in
IO format. The training data consisted of 185323 tokens while the test set contained 20592 tokens. For the training dataset, there are 8895 phrases, 3592 for LOC, 4071 for PER, and 1232 for ORG. In testing dataset, there are 634 instances of LOC, 107 of ORG, and 200 of PER. For training and predicting, we converted IO annotations to BIO / BIOES and added
sentence boundaries. A benchmarking of annotation scheme performance was conducted using Wapiti’s implementation of conditional random fields (CRF) \citep{lavergne-cappe-yvon:2010:ACL}\footnote{Available at \url{https://wapiti.limsi.fr/}}. Models were trained using the original corpus as well as a corpus with part of speech (POS) labels provided by TreeTagger. We converted all results to the IO annotation scheme to allow a fair comparison of the performance of each annotation method. Results were evaluated using the standard F1 metric as calculated by conlleval\footnote{Available at \url{https://www.clips.uantwerpen.be/conll2003/ner/}}. We find that the BIOES annotation scheme performs best, as it achieves the highest F1 score in both corpora. 

\begin{figure}[h]
\centering
\begin{tabular}{ c c c c }
& IO & BIO & BIOES \\ \hline
F1 score & 62.41 & 61.33 & 62.57
\end{tabular}
\caption{NER results using original corpus}
\end{figure}

\begin{figure}[h]
\centering
\begin{tabular}{ c c c c }
& IO & BIO & BIOES \\ \hline
F1 score & 63.39 & 64.58 & 65.62
\end{tabular}
\caption{NER results with POS tags}
\end{figure}

We employed the state-of-the-art transformer embeddings, Flaubert, with SequenceTagger from the Flair NLP library. The model utilized a bidirectional LSTM combined with word-level embeddings to handle contextual information. The new architecture improved the score from CRF to 79.12 with BIOES annotation, a significant jump from CRF, demonstrating the effectiveness of the approach using transformer-based embeddings and LSTM structure. 

LSTM outperforms CRF because it has the ability to capture long-term dependencies within a sequence, which is essential for NER because named entites can span multiple words. LSTM models utilize a memory cell to store information over time, allowing them to keep track of relevant context over longer sequences and better handle variations in word order and sentence structure. Additionally, pre-trained embeddings from Transformer models can capture rich contextual and semantic information about words. By incorporating these embeddings as input to LSTM layers, the combination of the two can reduce overfitting and improve generalization, resulting in even more accurate NER models.

We further improved the model's performance by applying two data augmentation techniques, label-wise token replacement and shuffle within segments, to the original training dataset with the BIOES annotation scheme. After applying these technqiues, the training dataset and entities increase about 2 times. The results indicated an increase in the F1 score for every category in the class, resulting in a final score of 79.39.

\begin{figure}[h]
\centering
\begin{tabular}{ c c c c }
& CRF  & BiLSTM + FlauBert & Data augmentatin \\ \hline
F1 score & 65.62 & 79.12 & 79.39
\end{tabular}
\caption{NER results with BIOES annotation}
\end{figure}

The results of the experiments suggest that the BIOES annotation scheme is superior to both the BIO and IO schemes in terms of F1 score for all label categories. The BIOES scheme is particularly effective at identifying precise entities and distinguishing boundaries between words, as evidenced by a 3-point increase in precision and recall for the ORG category. While precision is slightly decreased for other categories with BIOES, recall is higher, resulting in a slightly higher overall F1 score. This suggests that additional tags in the annotation process can help the model identify more correct phrases, but may also result in more false positives, thereby slightly reducing precision. Overall, these findings support the hypothesis that incorporating additional tags in the annotation process can lead to improved performance in named entity recognition tasks.

Interestingly, the data augmentation techniques were particularly effective for less commonly seen categories. Specifically, the augmentation increased F1 scores by less than 0.1 for LOC, 1 for PER, and almost 7 for ORG. The precision and recall increases for every category as well except a slightly decrease for recall for LOC category. The ORG category had the fewest instances in the training dataset, appearing only one sixth as often as the LOC category. These results demonstrate that data augmentation is especially useful for categories with fewer examples, as it creates more varied and diverse examples that enhance the model's learning. 

\section{Previous work}

\citet{ollagnier-EtAl:2014:TALN} used the Open Edition corpus the Quaero Broadcast News Extended Named Entity corpus\footnote{\url{http://catalog.elra.info/product_info.php?products_id=1195}}, which contains over 1.2M tokens.
They evaluated NER results with LIA\_NE (HMM-CRFs) \citep{frederic-charton:2010}, 
OpenNLP (Maximum entropy)\footnote{\url{https://opennlp.apache.org}} and 
Standford NER (CRFs)\footnote{\url{https://nlp.stanford.edu/software/CRF-NER.shtml}} with different sizes of training data. They found LIA\_NE performed the best, obtaining up to 57.9 F1 score. 
\citet{partalas-lopez-segond:2016:TALN} compared NER systems in the e-Commerce domain for cosmetics products by using handcrafted rules and machine learning techniques. 
They used two 50K tokens data sets (cosmetics magazines and blog articles) and concluded that a rule-based approach of a system of lexical combined syntactic rules with a domain-specific dictionary usually outperformed CRFs. 
Their rule-based systems yielded between 60.00 and 90.68 F1 scores based on different entities. 

There were efforts to create corpora annotated in named entities for French.
\citet{sagot-richard-stern:2012:TALN2012} and \citet{dutrey-EtAl:2012:TALN2012} manually annotated named entities in the French treebank, and in restricted domain such as oral dialogues recorded by the EDF call center for information extraction, respectively. 
\citet{okinina-EtAl:2013:TALN} enriched proper nouns by mining Wikipedia with the combination of DBpedia rules and a support vector machine classification. 
\citet{hatmi:2012:RECITAL2012} used a cross-lingual approach by converting a rule-based English NER system into French by using lexical and grammar adaptations. 

There are also applications of NER, where the use of NER systems improved the performance of other natural language processing tasks.
\citet{fraisse-paroubek-francopoulo:2013:TALN} employed NER for better classification results on opinion mining and sentiment analysis. 
\citet{sagot-gabor:2014:TALN} corrected OCRed named entities errors by using a rule-based NER system. 
\citet{brando-domingues-capeyron:2016} used NER for recognizing geographical references.
Additionally, \citet{dupont-tellier:2014:TALN} proposed a pipeline for French NER based on \texttt{Wapiti}. 

\citet{park:2018:TALN:NER} proposed different algorithms (HMM, CRFs, and bi-LSTM), semi-supervised learning and reranking approaches to improve French NER results up to a F1 score of 77.95, 34 points over baseline results. 
Since we used the same dataset that \citet{park:2018:TALN:NER} proposed, our results outperform the previous work using different annotation schemes and data augmentation. 

\section{Conclusion}
We believe that we achieved the best reported result for the current dataset. We find that the BIOES annotation scheme outperforms IO and BIO in French NER tasks. The use of transformer-based embeddings and LSTM structure was shown to improve model performance compared to CRFs. Furthermore, applying data augmentation techniques reduces overfitting and improves the generalizability of the model. The results of our experiments show that this outcome was especially noticeable in categories with fewer examples. This suggests that a consideration of extrinsic factors may improve performance in French NER tasks, particularly where data is limited. 

For future work, a comparison of other monolingual French and multilingual transformers may further improve results. Additionally, two other approaches for data augmentation, synonym replacement (SR) and mention replacement (MR), were not able to be conducted due to missing information in wordNET French. We hope to examine the effect of these two approaches when such data becomes available.








\end{document}